\documentclass[journal]{journal}
\ifCLASSINFOpdf
\else
\fi

\usepackage{hyperref}

\begin{document}
%
\title{A Generalized Weighted Loss for SVC and MLP}

%
%
%

\author{Filippo Portera}

%
%

\markboth{Journal of \LaTeX\ Class Files,~Vol.~6, No.~1, January~2007}%
{Shell \MakeLowercase{\textit{et al.}}: Bare Demo of IEEEtran.cls for Journals}
%



\maketitle
\thispagestyle{empty}

\begin{abstract}
Usually standard algorithms employ a loss where each error is the mere absolute difference between the true value and the extrapolation, in case of a regression task. In the present, we introduce several error weighting schemes that are a generalization of the consolidated routine.
We study both a binary classification model for Support Vector Classification and a regression net for Multy-layer Perceptron.
Results proves that the error is never worse than the standard procedure and several times it is better.
\end{abstract}

\begin{IEEEkeywords}
Machine Learning, Binary Classification, SVC, Regression, MLP.
\end{IEEEkeywords}

%
\IEEEpeerreviewmaketitle

\section{Introduction}
%
%
%
%
We would like to show that a standard loss generalization for binary classification (in our case we have chosen SVC and MLP) could produce results not worse w.r.t. the consolidated loss.
In fact, the possibility that a given data-set presents non-IID samples can be exploited by these generalized losses.

The loss studied to generalize SVC and the full optimization problem are:

\begin{equation}
P = \frac{1}{2}||\vec{v}||^2 + C \sum_{i=1}^l \xi_i w_i 
\end{equation}

suject to:

\begin{equation}
y_i (\vec{v}' \vec{x_i} + b) \geq 1 - \xi_i \hspace{1cm} i \in {1,..,l} 
\end{equation}

and:

\begin{equation}
  \xi_i \geq 0 \hspace{1cm} i \in {1,..,l} 
\end{equation}

where $\vec{v}$ represents the linear weights of the extraploator function, $l$ is the number of training examples, $C$ is a trade-off hyper-parameter, $\xi_i$ is the error on sample $i$, and $w_i$ are some scalar weights that are a function of a distribution $s_i$ of the samples:

\begin{equation}
s_i = \sum_{j=1}^{l} (e^{-\gamma_S ||\vec{x}_i - \vec{x}_j||^2}) 
\end{equation}
Other distribution can be adopted (e.g., 1 + the RBF norm instead of the RBF dot product). And let:
\begin{equation}
sy_i = y_i y_j \sum_{j=1}^{l} (e^{-\gamma_S||\vec{x}_i - \vec{x}_j||^2}) 
\end{equation}

with $\gamma_S$ additional hyper-parameter. Here lies the complexity of the algorithm since this calculation is $O(l^2)$. Perhaps it can be overtaken with pattern sampling or, in the case of MLP with a sort of weights learning.

\begin{equation}
w_i = f(s_i)    
\end{equation}

This implies a quadratic problem that is different from traditional SVC:

The Lagrangian would be:

\begin{equation}
  L = \frac{1}{2}||\vec{v}||^2 + C \sum_{i=1}^l \xi_i w_i - \sum_{i=1}^l \alpha_i ( y_i  (\vec{v}' \vec{x_i} + b) - 1 + \xi_i)
\end{equation}

\begin{equation}
  - \sum_{i=1}^l \eta_i \xi_i
\end{equation}

subject to:

\begin{equation}
  \alpha_i \geq 0 \hspace{1cm} i \in {1,..,l}
\end{equation}

\begin{equation}    
  \eta_i \geq 0 \hspace{1cm} i \in {1,..,l} 
\end{equation}

Applying the KKT condition for optimaility:

\begin{equation}
    \frac{\partial L}{\partial \vec{v}} = \vec{v} - \sum_{i=1}^1 \alpha_i y_i \vec{x_i} = 0 \Rightarrow \vec{v} = \sum_{i=1}^1 \alpha_i y_i \vec{x_i} 
\end{equation}

\begin{equation}
    \frac{\partial L}{\partial \vec{\xi}} = \vec{\alpha} + \vec{\eta} - C \vec{w} = 0\Rightarrow \vec{\alpha} \leq C \vec{w}
\end{equation}

\begin{equation}
    \frac{\partial L}{\partial b} = \sum_{i=1}^l \alpha_i y_i = 0
\end{equation}

Thus, the dual becomes:

\begin{equation}
 D = \sum_{i=1}^l \alpha_i - \frac{1}{2} \sum_{i=1}^l \sum_{j=1}^l \alpha_i y_i \alpha_j y_j K(\vec{x_i}, \vec{x_j})
\end{equation}

subject to:

\begin{equation}
  0 \leq \alpha_i \leq C w_i \hspace{1cm} i \in {1,..,l} 
\end{equation}

\begin{equation}
  \sum_{i=1}^l \alpha_i y_i = 0
\end{equation}

This is very similar to standard SVC dual \cite{Vapnik}, apart the constraints on the lagrangian multipliers.

We wrote an ad-hoc quadratic optimizer for this problem\footnote{The code of this work is available at \href{https://osf.io/yj8er/}{OSF GWL Project}}, with a SMO-like method (\cite{AiolliSperduti}).

We iteratively select 2 distinct multipliers and we modify them with an attempt to improve the dual objective function:

\begin{equation}
  \alpha_i^{t+1} = \alpha_i^{t} + \nu y_i
\end{equation}

\begin{equation}
  \alpha_j^{t+1} = \alpha_j^{t} - \nu y_j
\end{equation}

The motivation is the enforcement of the second dual constraint on the $  \sum_{i=1}^l \alpha_i y_i = 0$.

The $\nu$ in the optimal direction is obtained deriving $D$ by $\nu$ as it has been done in section 5.1 of (\cite{Portera}).

This direction is:

\begin{equation}
  \nu = \frac{y_j - y_i - \sum_{p=1}^l\alpha_p y_p K(\vec{x_j}, \vec{x_p}) + \sum_{p=1}^l\alpha_p y_p K(\vec{x_i}, \vec{x_p}) }{K(\vec{x_i},\vec{x_i}) - 2K(\vec{x_i},\vec{x_j}) + K(\vec{x_j},\vec{x_j})}
\end{equation}

Once the candidate $\nu$ has been determined, it has to be clipped in order to satisfy the constraints on both the multipliers.

At each iteration we compute $b$ with the suport vectors that lie in the margin (for which $0 < \alpha_i < C w_i$)
as it has been reported in \href{https://stats.stackexchange.com/questions/362046/how-is-bias-term-calculated-in-svms}{How to calculate $b$}.

The kernel used to compute $K(\vec{x}, \vec{y})$ is RBF with hyper-parameter $\gamma_K$.
The whole procedure is iterated $50 l^2$ times for each training problem.

\section{Related works}
 In (\cite{zhao}) they learn the loss weights directly from the training and validation sets. They assert that there is a substantial improvement in the generalization error and they also provide theoretical bounds.

\section{Method}
We use the acronym GWL for Generalized Weighted Loss.

We tried 4 distinct algorithms: the Python 3 package sklearn.svm.SVC, GWL SVC with $w_i = 1$, GWL (here we mean the generalized loss with $w_i$'s built as described), and GWL with random weights.
We would like to know if, in the general case, the optimal solutions use $w_i$ not equal to $1$.
We have selected at least 8 cases of study, to determine the weight $w_i$ of a sample $i$. Therefore, some evaluated weighting functions are:

\begin{enumerate}
\item 
\begin{equation}
w_i = \sqrt{s_i} 
\end{equation}
\item
\begin{equation}
w_i = s_i
\end{equation}
\item
\begin{equation}
w_i = s_i^2
\end{equation}
\item
\begin{equation}
w_i = \frac{1}{\sqrt{s_i}}
\end{equation}
\item
\begin{equation}
   w_i = \frac{1}{s_i}
\end{equation}
\item
\begin{equation}
   w_i = \frac{1}{s_i^2}
\end{equation}
\item
\begin{equation}
   w_i = sy_i
\end{equation}
\item
\begin{equation}
w_i = 1 + \textrm{rand}[0,1]
\end{equation}
\end{enumerate}

The case $8$ is useful to show that a weighting scheme based on the training distribution is more convenient w.r.t. a random weighting scheme.

\section{Results}
We explored a 2 dimensional hyper-parameters grid for sklearn.svm.svc, involving $\gamma_K$ and $C$.
While we used the additional hyper-parameter $\gamma_S$ to generete the loss weights.
That is the reason why the experiments with loss weights take more time to terminate.
Obviously, the second grid is an extension of (it covers) the first one.
Those are the results for the 5-fold cross-validation with data-sets extracted from the UCI website, and opportunely treated (double or inconsistent samples removed, shuffling):

\begin{table}
\begin{center}
\begin{tabular}{ |c|c|c|c| } 
 \hline
 Algorithm & Data-set & Mean F1 & Time \\ 
 \hline
sklearn.svm.SVC  &Ionosphere    &0.968638   &0m3,130s    \\
GWL SVC          &Ionosphere    &0.970651   &62m34,796s\\
GWL(1)           &Ionosphere    &0.977172   &175m44,060s\\
GWL(2)           &Ionosphere    &0.977172   &175m41,022s\\
GWL(3)           &Ionosphere    &0.977172   &187m1,032s\\
GWL(4)           &Ionosphere    &0.977172   &187m59,323s\\
GWL(5)           &Ionosphere    &0.977359   &3h:07m:41s\\
GWL(6)           &Ionosphere    &\textbf{0.977538}   &2h:58m:42s\\
GWL(8)           &Ionosphere    &0.977292   &3h:24m:32s\\   
GWL(8)           &Ionosphere    &0.977292   &3h:08m:31s  \\ 
GWL(8)           &Ionosphere    &0.974767   &3h:07m:36s    \\
GWL(8)           &Ionosphere    &0.975011   &3h:05m:40    \\
\hline
\end{tabular}
\caption{\label{tab_iono}SVC and GWL with Ionosphere data-set}
\end{center}
\end{table}

\begin{table}
\begin{center}
\begin{tabular}{ |c|c|c|c| } 
 \hline
 Algorithm & Data-set & Mean F1 & Time \\ 
 \hline
sklearn.svm.SVC  &Sonar         &0.886610   &0m2,396s\\
GWL SVC          &Sonar         &0.904489   &16m19,019s\\
GWL(1)           &Sonar         &0.909337   &28m8,391s\\
GWL(2)           &Sonar         &0.916513   &31m58,852s\\
GWL(3)           &Sonar         &0.908717   &36m26,726s\\
GWL(4)           &Sonar         &0.913580   &40m51,567s\\
GWL(5)           &Sonar         &0.916303   &45m46,03s\\
GWL(6)           &Sonar         &\textbf{0.916671}   &41m:17,92s\\
GWL(7)           &Sonar         &0.911098   &\\
GWL(8)           &Sonar         &0.907057   &      \\   
\hline
\end{tabular}
\caption{\label{tab_sonar}SVC and GWL with Sonar data-set}
\end{center}
\end{table}

\begin{table}
\begin{center}
\begin{tabular}{ |c|c|c|c| } 
 \hline
 Algorithm & Data-set & Mean F1 & Time \\ 
 \hline
sklearn.svm.SVC  &Breast        &0.959825   &0m3,387s\\
GWL SVC          &Breast        &0.958628   &174m47,138s\\
GWL(1)           &Breast        &0.963625   &432m3,673s\\
GWL(2)           &Breast        &0.963896   &448m14,994s\\
GWL(3)           &Breast        &\textbf{0.967909}   &443m59,346s\\
GWL(4)           &Breast        &0.966109   &401m18,642s\\
GWL(5)           &Breast        &0.964666   &6h:48m:01s\\
GWL(6)           &Breast        &0.964666   &6h:34m:13s\\
GWL(8)           &Breast        &0.961837   &8h:02m:57s \\ 
\hline
\end{tabular}
\caption{\label{tab_breast}SVC and GWL with Breast data-set}
\end{center}
\end{table}

\begin{table}
\begin{center}
\begin{tabular}{ |c|c|c|c| } 
 \hline
 Algorithm & Data-set & Mean F1 & Time \\ 
 \hline
sklearn.svm.SVC  &Statlog       &0.610351   &0m38,297s\\
GWL SVC          &Statlog       &0.644108   &1725m29,269s\\
GWL(1)           &Statlog       &0.651278   &5370m49,360s\\
GWL(2)           &Statlog       &0.651278   &5499m18,497s\\
GWL(3)           &Statlog       &0.644329   &5638m2,544s\\
GWL(4)           &Statlog       &0.649529   &5692m55,211s\\
GWL(5)           &Statlog       &0.651947   &5351m40,975s\\
GWL(6)           &Statlog       &\textbf{0.652786} &5546m26,466s\\
GWL(8)           &Statlog       &0.646312   &83h:50m:08s\\
\hline
\end{tabular}
\caption{\label{tab_statlog}SVC and GWL with Statlog data-set}
\end{center}
\end{table}

We also have tried 2 MLP nets with PyTorch on a regression task with $w_i = s_i$ and results are interesting (but the random initialization of the net weights should be considered in this case: Wine data-set: 3961 samples, 11 features; MLP: 100, 50, 20, 1, and Wine data-set, different MLP 100, 80, 40, 1 nodes per layer.
The theory underneath deep neural architectures can be foun in \cite{GoodfellowBengioCourville}.

\begin{table}
\begin{center}
\begin{tabular}{ |c|c|c| } 
 \hline
 Standard MLP MAE & gamma Best & Best Loss MLP MAE \\ 
 \hline
0.5871212    &              10      &    \textbf{0.57575756} (1)  \\
0.5694444    &              10      &    \textbf{0.5530303}  (2)  \\
0.5568182    &               1      &    \textbf{0.53661615} (3)  \\
0.5580808    &               0.01    &   \textbf{0.540404}   (4)  \\
0.54924244    &              0.1     &   \textbf{0.54671717} (5)  \\
0.510101     &             100       &   0.510101   (6)  \\
\hline
\end{tabular}
\caption{\label{tab_FirstMLP}First MLP with wine data-set}
\end{center}
\end{table}

\begin{table}
\begin{center}
\begin{tabular}{ |c|c|c| } 
 \hline
 Standard MLP MAE & gamma Best & Best Loss MLP MAE \\ 
 \hline
0.5694444      &            10      &    \textbf{0.5580808}  (1)  \\
0.53661615      &            1      &    \textbf{0.5290404}  (2)  \\
0.5378788       &          100      &    0.5378788  (3) \\
0.5770202       &           10      &    \textbf{0.5580808}  (4)  \\
0.54924244      &            0.01   &    \textbf{0.5252525}  (5)  \\
0.56565654      &            1      &    \textbf{0.5555556}  (6)  \\
\hline
\end{tabular}
\caption{\label{tab_SecondMLP}Second MLP with wine data-set}
\end{center}
\end{table}

In this scenario it would be useful to determine the difference between eah couple of values, to understand which is the strategy that, in the most of the cases, performs best with the test set.
An idea is to learn weights, starting to run in parallel $n$ nets with different random weight vectors and selecting at each parallel one the best vector in terms of MAE and perturbing it and re-run the procedure for a given amount of iterations.

GWL has been written in C.
The regression code for the wine data-set has been written in Python 3.10 and torch.

Hardware employed: a notebook with 8 cores Intel(R) i5-10210U CPU @ 1.60GHz and 16GB of RAM,
and a PC with 16 cores 11th Gen Intel(R) Core(TM) i7-11700 @ 2.50GHz and 32 GB of RAM.

Baseline SVC algorithms have been measured on the notebook, while GWL times have been determined with the PC.

\section{Conclusion}
Results confirm the theory, they're not worse than the particular case.
In particular, it looks like that the preferred generalization scheme is the one that gives more importance to patterns that are isolated, on 3 data-sets from 4 for the SVC case.
Nevertheless it should be considered the fact concerning the unique geometry of each data-set, so each generalization scheme should be tested. 
The next step would be to leverage this method in order to learn the weights.
Perhaps this generalization could be employed in other contests such as SVR, multi-class classification, and other MLPs.


%

\ifCLASSOPTIONcaptionsoff
  \newpage
\fi


\begin{thebibliography}{1}

\bibitem{zhao}
  "Zhao Sen et al.", \emph{Metric-Optimized Example Weights}, 2019.

\bibitem{AiolliSperduti}
  Aiolli F, Sperduti A., \emph{An efficient SMO-like algorithm for multiclass SVM},
  Proceedings of the 12th IEEE Workshop on Neural Networks for Signal Processing,
  pp, 297--306, 2002/9/6

\bibitem{GoodfellowBengioCourville}
  Book: \emph{Deep Learning},
  MIT Press, 2016

\bibitem{Vapnik}
  Book: \emph{Statistical Learning Theory}
  WILEY, 1998

\bibitem{Portera}
  Portera F., \emph{A generalized quadratic loss for SVM and Deep Neural Networks}
  LOD 2020 Conference work

\end{thebibliography}
\end{document}